\documentclass[runningheads]{llncs}

\usepackage{eccv}

\usepackage{eccvabbrv}

\usepackage{multirow}
\usepackage{graphicx}
\usepackage{booktabs}
\usepackage{bbding}
\usepackage{wrapfig}
\usepackage{placeins}

\usepackage[accsupp]{axessibility}  % Improves PDF readability for those with disabilities.

\usepackage{hyperref}
\usepackage[normalem]{ulem}
\usepackage{orcidlink}

\newcommand{\ourmodel}[1]{\textcolor{orange}{\{ModelName\}}}
\newcommand{\ourdataset}[1]{\textcolor{blue}{\{DatasetName\}}}
\newcommand{\ourbenchmark}[1]{\textcolor{green}{\{BenchmarkName\}}}

\usepackage[dvipsnames]{xcolor}
\usepackage{pifont}
\begin{document}

\title{Extending a Large View Synthesis Model for Multi-view Panoptic Segmentation} 

\titlerunning{Extending LVSM for Panoptic Seg}

\author{Kwonyoung Ryu\inst{1}\orcidlink{0000-0002-9685-5155}, In-Jae Lee\inst{4}\orcidlink{0009-0004-3654-8614}, Jonghyun Jin\inst{2}\orcidlink{0009-0007-8611-8665}, Hyunjee Lee\inst{2}\orcidlink{0009-0002-3537-5673}, \\Jongmin Lee\inst{3}\orcidlink{0000-0001-9410-027X}, Jaesik Park\inst{4}\orcidlink{0000-0001-5541-409X}}

\authorrunning{K. Ryu et al.}

\institute{$^{1}$POSTECH \quad $^{2}$POSCO DX \quad $^{3}$Chung-Ang University \quad $^{4}$Seoul National University}

%\institute{POSTECH \and Seoul National University \and POSCO DX \and Chung-Ang University}

\maketitle

\begin{abstract}
  Large view synthesis models synthesize novel views through cross-view attention without explicit 3D representations, and recent studies have shown that they learn accurate spatial correspondence from RGB supervision alone. 
We observe that this correspondence generalizes beyond appearance. 
When non-photorealistic signals such as binary encoded panoptic labels are passed through the model, they are propagated to novel views with consistent spatial structure. 
These results indicate that the correspondence learned for RGB view synthesis can also propagate view-independent per-pixel labels.
From this observation, we present the first work to extend large view synthesis models beyond appearance rendering to 3D scene understanding. 
We propose a panoptic segmentation pipeline that reuses a frozen view synthesis model to propagate panoptic labels from input views to novel views, without 3D reconstruction or any segmentation-specific training of the view synthesis model. 
Given panoptic labels on the input views, we encode them into binary channel representations and pass them through the same model to render target-view segmentation. 
On ScanNet, our method achieves segmentation quality on par with Gaussian based approaches requiring explicit 3D reconstruction, while outperforming them in novel view synthesis by more than 7 dB. 
The label propagation also transfers across datasets, surpassing these approaches on Replica without any fine-tuning.

\end{abstract}
\section{Introduction}

\begin{figure*}[t]
    \centering
    \includegraphics[width=\textwidth]{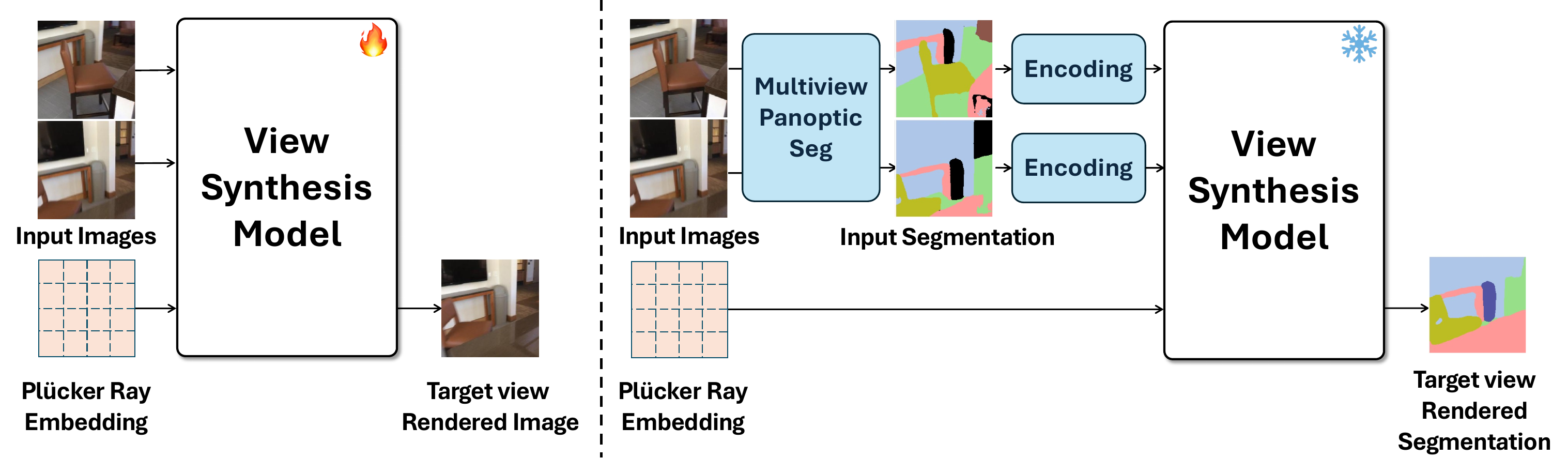}
    
    \caption{
        \textbf{Illustration of a large-view synthesis model and its extension to multi-view panoptic segmentation.}
        In the \textbf{left} figure, a feed-forward large view synthesis model renders a novel target image from the input source views.
        In the \textbf{right} figure, the same model is bootstrapped for panoptic segmentation. 
        We propose a simple signal-level extension that repurposes the learned cross-view correspondence of a view synthesis model for 3D scene understanding beyond RGB rendering.}
    \label{fig:teaser}
\end{figure*}

% P1: Task motivation + explicit 3D 
Predicting panoptic segmentation from novel viewpoints requires assigning semantic labels and instance identities to every pixel in an unobserved view, given only a few images of the scene. 
Embodied agents need this capability to reason about objects beyond the currently observed view. 
Example applications include anticipating scene content before navigation and augmenting training data from a small number of annotated views.

Most existing approaches first reconstruct a 3D representation such as NeRF~\cite{mildenhall2020nerf} or 3D Gaussian Splatting~\cite{3dgs23}, and then lift 2D panoptic labels into the recovered representation through per-scene optimization~\cite{panopticlifting23, n2f2}.
This design couples reconstruction quality with segmentation quality. Errors in the recovered geometry propagate directly to the lifted labels.
Recent feed-forward variants predict 3D Gaussians with semantic features in a single forward pass~\cite{lsm24, SIU3R}, eliminating per-scene optimization but still coupling reconstruction quality with segmentation in a shared 3D representation.

% P2: View synthesis models - different paradigm + unexplored territory
These limitations motivate an alternative that avoids explicit 3D reconstruction entirely.
Meanwhile, recent NVS methods replace explicit 3D representations with latent token-based scene modeling~\cite{srt, rust23, lvsm, rayzer, less3depend}.
These large view synthesis models attend across source and target tokens to synthesize novel views directly, without any 3D inductive bias such
as NeRF~\cite{mildenhall2020nerf} or 3DGS~\cite{3dgs23} in the network architecture~\cite{lvsm}.
Recent studies~\cite{less3depend, rayzer} show that these models learn cross-view spatial correspondence even without explicit 3D supervision, and that their attention layers learn accurate cross-view spatial correspondence between source and target views.
However, all existing analyses focus exclusively on target RGB view appearance. Whether the learned implicit correspondence extends to signals beyond appearance and can thus serve tasks beyond view synthesis remains unexplored.

% P3: What we've found
We investigate this question by analyzing the implicit correspondence of a large view synthesis model~\cite{less3depend} fine-tuned only for target RGB reconstruction via gradient-based saliency analysis~\cite{deconv14}. We find that the model attends to geometrically corresponding regions in source views regardless of whether the input is RGB format or non-photorealistic per-pixel label encodings.
The model thus learns geometric relationships between views rather than appearance-specific features, and arbitrary per-pixel signals can transfer to novel views through the same large view synthesis model.

% P4: 접근법 + 결과
We present the first work to extend large view synthesis models beyond appearance rendering to 3D scene understanding. We propose a panoptic segmentation pipeline that bootstraps a large view synthesis model~\cite{less3depend} trained solely for RGB image reconstruction.
We first segment the input views using a shared query decoder to obtain cross-view-consistent panoptic labels, because the propagation stage preserves the labels provided as input, and instance identities must be consistent before encoding.
The target RGB rendering path reconstructs the novel target-view image from source views, and the segmentation path encodes the panoptic labels as binary channel encodings and passes them through the same novel view synthesis model to produce target-view segmentation.
Our method achieves 33.56 PSNR and 0.5949 mIoU on novel views in the ScanNet dataset~\cite{scannet}, higher than the baseline SIU3R~\cite{SIU3R} (25.88 PSNR, 0.5894 mIoU), which trains explicit 3D Gaussians with dedicated segmentation objectives.
Because the rendering transformer remains frozen, our pipeline preserves full rendering quality, exceeding Gaussian-based methods by more than 7~dB.

% P5: Contributions
Our contributions are as follows.
\begin{itemize}
    \item To our knowledge, this is the first work to extend large view synthesis models beyond appearance rendering to 3D scene understanding, demonstrating that explicit 3D reconstruction is not a prerequisite for multi-view panoptic segmentation.
    \item We demonstrate, through gradient-based saliency analysis, that a large view synthesis model trained only for RGB reconstruction propagates panoptic labels to novel views. The view synthesis model receives no segmentation-specific supervision.
    \item We present a modular pipeline that decouples novel view rendering and segmentation, allowing independent replacement of either component as the respective fields advance.
\end{itemize}

\section{Related Work}

\noindent\textbf{Scene Understanding and Reconstruction.}
Panoptic segmentation in 3D assigns semantic labels and instance identities to every element of a scene.
Some methods directly segment pre-scanned point clouds~\cite{mask3d23, oneformer3d24, openscene23}, but they assume the 3D geometry is already available.
When only images are given, the dominant paradigm lifts 2D predictions into a 3D representation through per-scene optimization.
NeRF-based approaches~\cite{semanticnerf21, panopticnerf22, pnf22,
panopticlifting23, contrastivelift24} fuse noisy 2D labels into a consistent volumetric model, while Gaussian Splatting variants~\cite{3dgs23} embed semantic or instance features directly into primitives~\cite{feature-3dgs, langsplat, gaussian_grouping, opengaussian24, fmgs25, plgs24, pcflift24}.
A related direction distills open-vocabulary features from 2D foundation models into radiance fields~\cite{lerf23, n2f2, clipgs24}.
All of these approaches require posed images, dense captures, and minutes to hours of optimization per scene.
Recent feed-forward methods remove per-scene optimization by predicting labeled 3D representations in a single forward pass without pose information~\cite{lsm24, SIU3R, panst3r25}.
However, they still depend on explicit 3D intermediates such as pixel-aligned Gaussians~\cite{lsm24, SIU3R} or dense point maps~\cite{panst3r25}.
As a result, reconstruction quality directly constrains segmentation quality, and the two components cannot be upgraded independently.
Our approach passes encoded panoptic labels through a large view synthesis model that has never been trained on segmentation.
This requires no 3D reconstruction, no task-specific training, and no architectural modification.

\noindent\textbf{Large View Synthesis Models.}
Novel view synthesis has traditionally relied on explicit 3D representations optimized per scene, such as volumetric fields~\cite{mildenhall2020nerf, mipnerf21, instantngp22, zipnerf23} or Gaussian primitives~\cite{3dgs23, mipsplatting24, 2dgs24}.
Feed-forward variants predict these representations from sparse inputs in a single pass, with some requiring posed images~\cite{pixelnerf21, ibrnet21, pixelsplat, mvsplat, lrm23, gslrm24} and others jointly recovering geometry and cameras from unposed images~\cite{dust3r, mast3r, noposplat, splatt3r24, vggt25}.
Despite their diversity, all assume that an explicit geometric intermediate must be recovered before rendering.

Another line of work synthesizes views directly through learned token-to-token mappings without explicit 3D structure~\cite{srt, osrt22, lvsm, rust23, rayzer, less3depend}.
Early models in this line require ground-truth camera poses~\cite{srt, lvsm}, but subsequent work progressively removes this dependency through self-supervised pose prediction~\cite{rayzer} and by eliminating pose representations entirely~\cite{less3depend}. 
These models use cross-view attention to model correspondence between source and target views, and recent analyses~\cite{less3depend, rayzer} show that this correspondence reflects 3D scene structure.
Yet they have been explored exclusively for appearance synthesis. Whether the implicit correspondence extends to other per-pixel signals has not been investigated.
We build upon Less3Depend~\cite{less3depend} and show through gradient attribution analysis that its learned implicit correspondence is input-agnostic, propagating panoptic labels to novel views without any task-specific adaptation.

\noindent\textbf{2D Panoptic Segmentation.}
Our pipeline uses off-the-shelf 2D panoptic segmentation as a modular source-view component.
The task was introduced as a unification of semantic and instance segmentation~\cite{panopticseg19}, with early solutions building on detection~\cite{maskrcnn17} or encoder-decoder~\cite{deeplab15} architectures.
The dominant modern formulation treats panoptic segmentation as mask classification via transformer
decoders~\cite{maskformer21, mask2former22, oneformer23}, and has been further extended toward open-vocabulary capabilities~\cite{sam23} and unified multi-task frameworks~\cite{oneformer23}.
In parallel, self-supervised vision transformers~\cite{dino21, dinov2} serve as dense-prediction backbones, bridged to segmentation heads via adapter modules~\cite{vitadapter23}.
We combine these components to extract panoptic labels from each input view and propagate them through the large view synthesis model.
Since no part of our pipeline is trained on 3D data, the 2D segmentation model can be replaced with any improved future method without modifying the rest of the system.
\section{Method}
\label{sec:method}

In this section, we briefly review large view synthesis models (Sec.~\ref{sec:preliminaries}) and their inherent implicit correspondence capability (Sec.~\ref{sec:implicit_corres}). We then present our approach to bootstrapping the view synthesis model for multi-view panoptic segmentation (Sec.~\ref{sec:bootstrap_nvs}), introduce a panoptic encoding and decoding scheme (Sec.~\ref{sec:seg_enc_dec}), and discuss training (Sec.~\ref{sec:training}) and design properties (Sec.~\ref{sec:discussion}).

\subsection{Preliminaries : Large View Synthesis Models}
\label{sec:preliminaries}

Novel view synthesis aims to render images from unobserved viewpoints given a set of source images.
Given $N$ source views $\{I^\mathrm{s}_i\}_{i=1}^{N}$ with $I^\mathrm{s}_i \in \mathbb{R}^{H \times W \times 3}$, the goal is to produce a target image $\hat{I}^\mathrm{t} \in \mathbb{R}^{H \times W \times 3}$ at a novel viewpoint.
Recent large view synthesis models such as Less3Depend~\cite{less3depend} and RayZer~\cite{rayzer} achieve strong performance even in the unposed sparse-view setting, where neither source nor target camera poses are provided. These models take $N$ source images as input and directly synthesize the target view by inferring relative geometry internally, without constructing any explicit 3D representation. 
In this paper, we adopt Less3Depend~\cite{less3depend}, which builds upon LVSM~\cite{lvsm}, a transformer-based renderer that tokenizes source views via a frozen DINOv2~\cite{dinov2} encoder and synthesizes the target view through joint attention across all source and target tokens. Unlike LVSM, Less3Depend~\cite{less3depend} does not require camera poses, enabling novel view synthesis from unposed sparse views. This formulation exposes a learned source-to-target correspondence through attention, which we later reuse to propagate panoptic labels without modifying the renderer. We examine whether this correspondence can also transfer non-RGB signals across views.

\begin{figure*}[t]
    \centering
    \includegraphics[width=\textwidth]{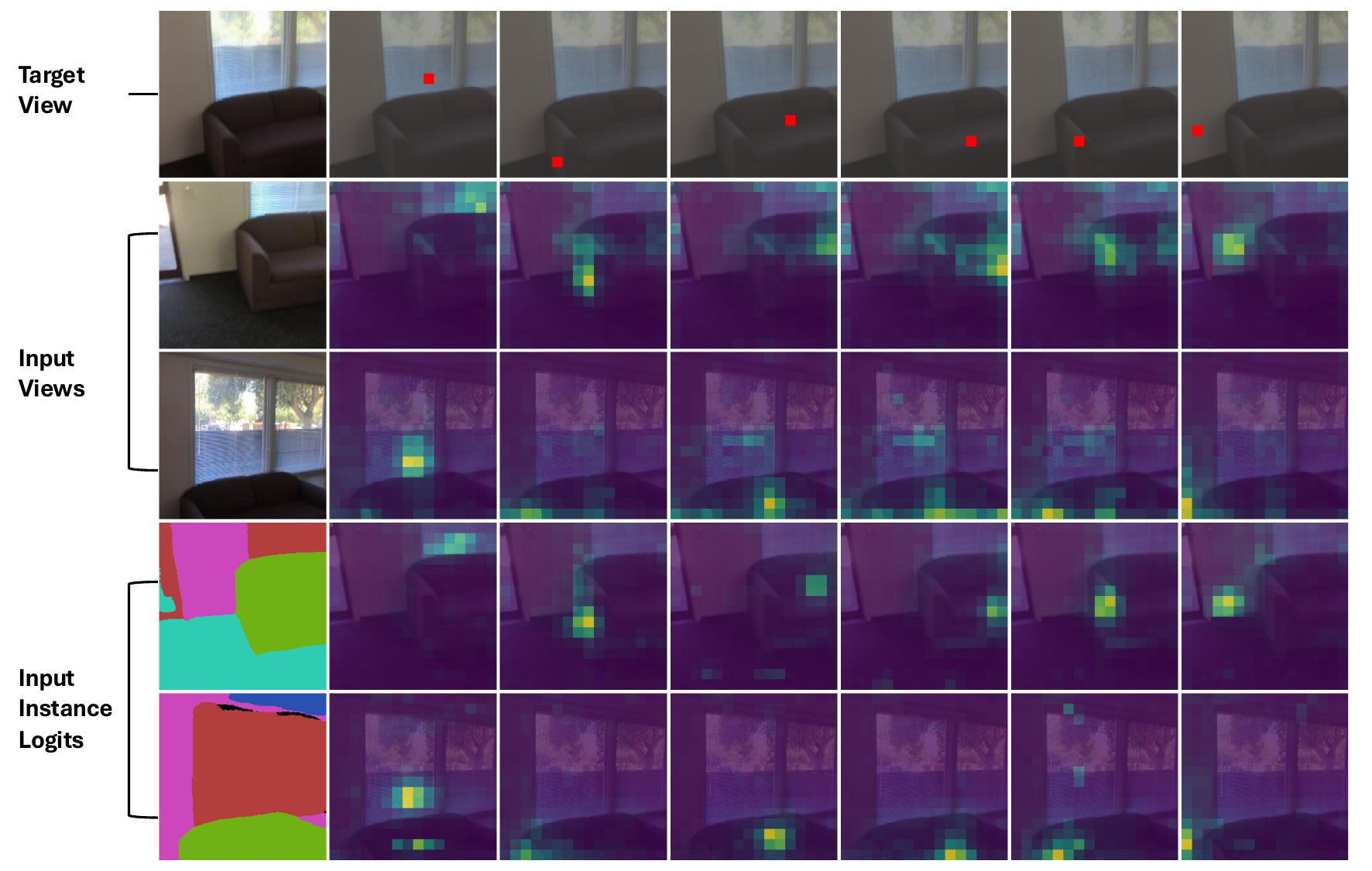}
    \caption{
        \textbf{Visualization of gradient saliency from target to source views.} 
        The saliency maps highlight similar source regions whether the model receives RGB images or binary instance encodings. This indicates that the learned cross-view correspondence of the model is not limited to appearance.
        Row 1 shows the rendered target view, with each column indicating the location of a different query patch in red. For each query patch, Rows 2-3 and Rows 4-5 show the gradient saliency over the same two source views, using RGB inputs and binary instance encodings, respectively.
    }
    \label{fig:attention}
\end{figure*}

\subsection{Implicit Correspondence in a Large View Synthesis Model}
\label{sec:implicit_corres}

In this section, we analyze whether the geometric correspondence learned by feed-forward large view synthesis models generalizes beyond RGB inputs, and show that this property enables panoptic label propagation to novel views.
To examine whether this correspondence is specific to RGB inputs, we visualize gradient saliency~\cite{deconv14} of the large view synthesis model~\cite{less3depend}. For a single target patch, we backpropagate through the decoder and encoder and record the absolute gradient magnitude at each source token. This directly measures how much each source region causally influences the target output. For both input types, we provide the target-view latent Pl\"ucker estimated from RGB inputs.

As shown in Fig.~\ref{fig:attention}, the gradient saliency reveals spatial correspondence between the target and source views. For each query patch on the target view, the saliency concentrates on geometrically corresponding regions in the source views (Rows~2,3). To test whether this correspondence depends on the input modality, we replace the source RGB inputs with binary instance encodings while keeping the same pose conditioning from the RGB rendering path. The correspondence persists under this change (Rows~4,5). The consistent saliency patterns across both modalities indicate that the model resolves spatial correspondence from geometric pose rather than input content, which enables panoptic label propagation through the rendering path.

\subsection{Bootstrapping NVS for Panoptic Understanding}\label{sec:bootstrap_nvs}

\begin{figure*}[t]
    \centering
    \includegraphics[width=\textwidth]{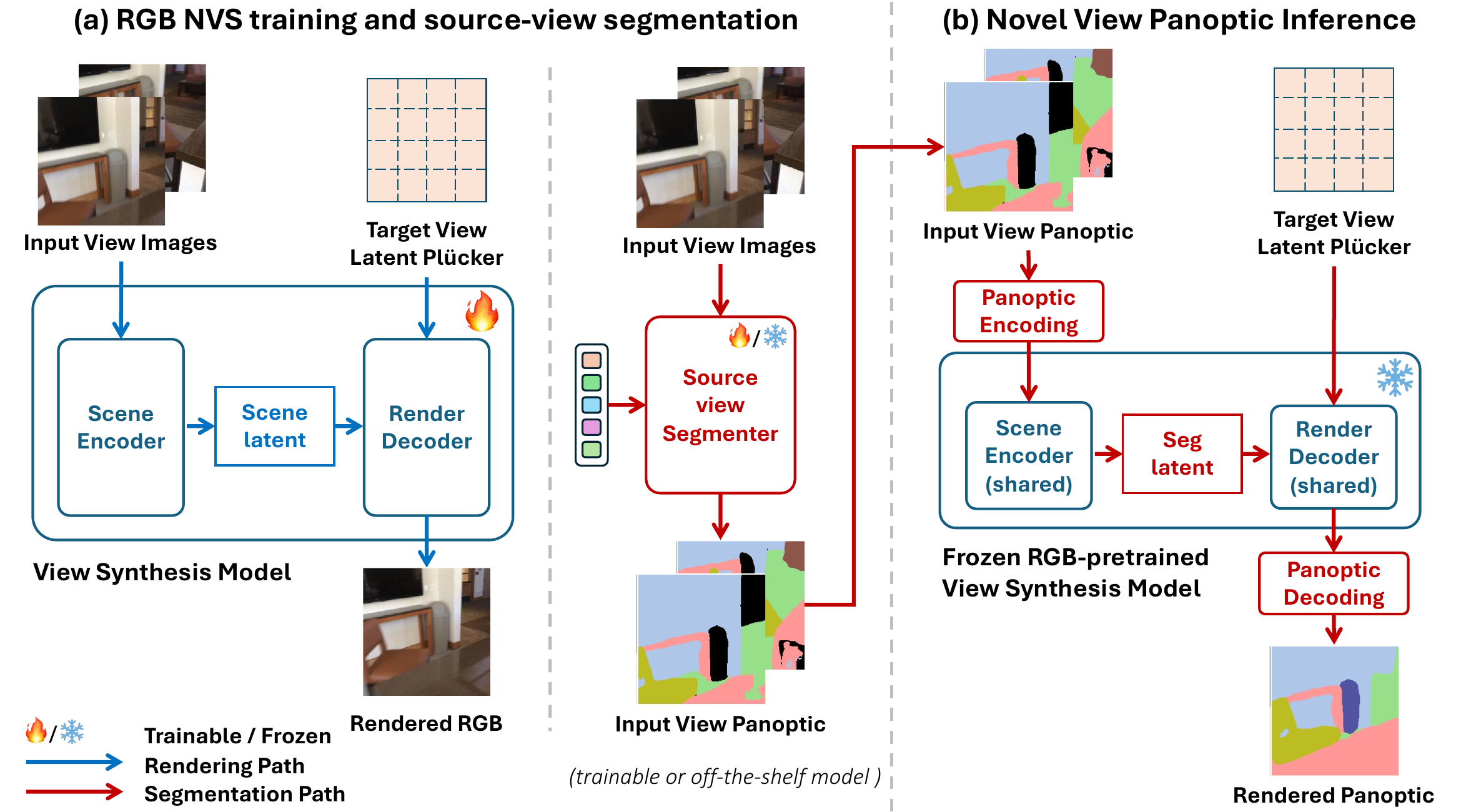} 
    \caption{
    \textbf{Overall pipeline.}
        (a) RGB NVS training and source-view segmentation: the view synthesis model (blue) is trained to encode input-view images into a scene latent and render the target-view RGB image from the target-view latent Pl\"ucker embedding. A source-view segmenter (red), implemented either as our trainable shared query decoder or an off-the-shelf model, predicts panoptic labels for the input views.
        (b) Novel-view panoptic inference: the input-view panoptic labels produced in (a) are encoded and passed through the frozen RGB-pretrained view synthesis model to propagate them to the target view. The resulting target-view binary output is converted into the final panoptic map by panoptic decoding.
        The view synthesis model is trained only for novel-view RGB synthesis, not for semantic or panoptic segmentation.
    }
    \label{fig:pipeline}
\end{figure*}

\noindent\textbf{Overall Pipeline.}
As illustrated in Fig.~\ref{fig:pipeline}(a), our method operates in two parallel paths through the same large view synthesis model.
Given $N$ unposed input views $\{I^\text{s}_i\}_{i=1}^{N}$ with 
$I^\mathrm{s}_i \in \mathbb{R}^{H \times W \times 3}$ and a target image $I^\mathrm{t}$, 
the model first encodes the input views into a scene latent $\mathbf{z}$ and estimates the target camera embedding:
\begin{equation}
  \mathbf{z} = \mathcal{E}(I^\mathrm{s}_1, \ldots, I^\mathrm{s}_N, \mathbf{z}_0),  \quad 
  \hat{\mathbf{e}}^\mathrm{t} = \mathcal{P}(I^\mathrm{t}, \; \mathbf{z}),
\end{equation}
where $\mathbf{z}_0$ is a learnable latent refined by 
attending to source-view tokens, $\mathcal{E}$ is the scene encoder, $\mathcal{R}$ is the render decoder, and $\hat{\mathbf{e}}^\mathrm{t}$ is a latent Pl\"ucker embedding estimated by the pose estimator $\mathcal{P}$ from the unposed target image $I^\mathrm{t}$ and the scene latent $\mathbf{z}$ following~\cite{less3depend}.
The rendering path synthesizes the target-view image through the decoding stage as follows,
\begin{equation}
  \hat{I}^\mathrm{t} = \mathcal{R}(\mathbf{z}, \; \hat{\mathbf{e}}^\mathrm{t}).
\end{equation}

For the segmentation path, a shared query decoder $\mathcal{D}$ first extracts cross-view consistent panoptic labels from the input views. As shown in Fig.~\ref{fig:pipeline}(b), the labels are then encoded into binary channel representations 
$\mathbf{b}_i \in \{0,1\}^{H \times W \times 3}$.
These encodings are then passed through the same encoder and decoder as follows.
\begin{equation}
  \mathbf{z}^{\mathrm{seg}} = \mathcal{E}(\mathbf{b}^\mathrm{s}_1, \ldots, \mathbf{b}^\mathrm{s}_N, \mathbf{z}_0), \quad
  \hat{\mathbf{b}}^\mathrm{t} = \mathcal{R}(\mathbf{z}^{\mathrm{seg}}, \; \hat{\mathbf{e}}^\mathrm{t}),
\end{equation}
where $\hat{\mathbf{e}}^\mathrm{t}$ and $\mathbf{z}_0$ are reused from the rendering path and all weights of $\mathcal{E}$ and $\mathcal{R}$ are trained with RGB reconstruction only.
The continuous output $\hat{\mathbf{b}}^\mathrm{t} \in [0,1]^{H \times W \times 3}$ is then decoded into the final panoptic map $P^\mathrm{t}$.
As shown in Sec.~\ref{sec:implicit_corres}, the correspondence learned by the view synthesis model remains stable when RGB inputs are replaced with binary instance encodings.
Since the model is trained solely for RGB image reconstruction, panoptic propagation requires only lightweight processing around the existing rendering path.

\noindent\textbf{Multi-view Panoptic Segmentation.}
Since our pipeline propagates input-view panoptic labels through the large view synthesis model, instance identities must be consistent across all input views. Conflicting labels for the same object would lead to ambiguous propagation.
To this end, we adopt a shared query decoder $\mathcal{D}$ following 
Mask2Former~\cite{mask2former22}, built on a frozen DINOv2~\cite{dinov2} 
encoder with ViT-Adapter~\cite{vitadapter23}.
A shared set of learnable object queries $\mathbf{Q}$ is fed into the decoder 
along with features from all input views simultaneously as follows.
\begin{equation}
  \{(\mathbf{m}_j, \, c_j)\}_{j=1}^{Q} = \mathcal{D}(\mathbf{Q}, \; I_1, \ldots, I_N),
\end{equation}
where $\mathbf{Q} = \{\mathbf{q}_j\}_{j=1}^{Q}$ is a set of $Q$ learnable object queries, $\mathbf{m}_j \in [0,1]^{N \times H \times W}$ is the predicted mask, and $c_j \in \mathcal{C}$ is the predicted semantic class for the $j$-th query.
Since each query attends to all views jointly, it explicitly represents a potential object instance across views, ensuring consistent instance identities without requiring explicit cross-view matching. 

\subsection{Panoptic Encoding and Decoding}\label{sec:seg_enc_dec}

The large view synthesis model transfers implicit correspondence through three output channels.
Given a panoptic map with semantic labels $S(p)$ and instance IDs 
$I(p) \in \{0, \ldots, K{-}1\}$ for each pixel $p$, we encode each 
pixel as $\mathbf{b}(p) = \mathcal{B}(I(p)) \in \{0,1\}^3$, where 
$\mathcal{B}$ denotes the 3-bit binary encoding ($K \leq 8$).
We also store a look-up table $\mathcal{L}$ mapping each instance index 
to its semantic class.
At inference, the model outputs 
$\hat{\mathbf{b}}^\mathrm{t} \in [0,1]^{H \times W \times 3}$ for the target view, 
and we recover the panoptic map as,
\begin{equation}
  \hat{I}^\mathrm{t}(p) = \mathcal{B}^{-1}\bigl(\mathrm{round}(\hat{\mathbf{b}}^\mathrm{t}(p))\bigr), 
  \quad S^\mathrm{t}(p) = \mathcal{L}\bigl(\hat{I}^\mathrm{t}(p)\bigr).
\end{equation}
Pixels where any channel $k$ satisfies $|\hat{b}_{:, :, k}(p) - 0.5| \leq \tau$ are marked as uncertain and left unlabeled. We set $\tau = 0.2$ for the best performance.

We choose binary over continuous encoding for robustness to rendering artifacts.
When the model blends two binary codewords at instance boundaries, 
the interpolated values converge toward $0.5$, which is maximally distant from both valid states and reliably rejected by thresholding.
A finer encoding would narrow the inter-codeword gaps, causing boundary interpolation to alias into valid but incorrect codewords and introduce phantom instances that cannot be filtered out.
A single 3-bit propagation pass represents at most eight active instance IDs. We handle scenes with more instances through a multi-pass decoding scheme, described in the supplementary (Appendix.\textcolor{red}{B}).

\subsection{Training Scheme}\label{sec:training}

\noindent The default variant fine-tunes the NVS model ($\mathcal{E}$, $\mathcal{P}$, $\mathcal{R}$) with the photometric loss of Less3Depend~\cite{less3depend}, using RGB reconstruction only. We train the shared query decoder $\mathcal{D}$ with the standard Mask2Former~\cite{mask2former22} objective on the input-view labels. The NVS model receives no segmentation-specific loss, so panoptic propagation relies entirely on the correspondence learned from photometric supervision. The segmentation path runs only a forward pass through $\mathcal{E}$ and $\mathcal{R}$, and no gradient flows into the rendering weights from the binary encodings. This keeps the rendering transformer identical to the frozen view synthesis model. The modular variants take source-view labels from an off-the-shelf segmenter such as SAM2~\cite{sam2} or PanSt3R~\cite{panst3r25}, and require no training within our pipeline. The propagation also holds when we replace Less3Depend~\cite{less3depend} with RayZer~\cite{rayzer} as the NVS backbone, showing that it applies across different large view synthesis models. We report the segmentation loss terms and weights, together with the RayZer results, in the supplementary (Appendix.\textcolor{red}{A} and \textcolor{red}{D}).

\subsection{Discussion}\label{sec:discussion}

\noindent A key reason the label propagation works is that panoptic labels are view-independent. The semantic class and instance identity of a surface point do not change with the viewpoint of the observer. The model only needs to establish accurate spatial correspondence between views, and the labels transfer directly.
This design separates source-view segmentation from target-view propagation. The large view synthesis model and the source-view segmenter can be replaced without changing the propagation stage, without retraining the rest of the pipeline. Replacing the source-view segmenter with PanSt3R~\cite{panst3r25} transfers to Replica~\cite{StraubX19Replica} zero-shot and outperforms a reconstruction-based baseline on every metric, without any fine-tuning of the rest of the pipeline (Sec.~\ref{sec:modular_replica}).
\section{Experiments}

\noindent In this section, we evaluate rendering quality, novel-view segmentation accuracy, low-overlap behavior, and cross-dataset transfer. We evaluate whether panoptic propagation through a frozen NVS model preserves rendering quality while reaching competitive segmentation accuracy, under standard overlap (Sec.~\ref{sec:main_results}) and sparse overlap (Sec.~\ref{sec:low_overlap}). We compare design alternatives that couple segmentation with the NVS model (Sec.~\ref{sec:design_alternatives}). We then assess the decoupled design through modular composition with an off-the-shelf segmenter and zero-shot transfer to a new dataset (Sec.~\ref{sec:modular_replica}).

\subsection{Experimental Setup}\label{sec:experimetal_setup}

\noindent\textbf{Implementation Details.}
We use ScanNet~\cite{scannet} for training and evaluation. Each training sample consists of 2 input views and 2 target novel views, all resized to $224 \times 224$. Following SIU3R~\cite{SIU3R}, view pairs are sampled with depth-based IoU in $[0.3, 0.8]$ for both training and evaluation. For evaluation, we use 2 input views and 4 target views on 1,860 selected view pairs in the same IoU range from the validation set. We train for 100 epochs on 8 NVIDIA RTX A6000 GPUs with a total batch size of 64, taking approximately 3 hours.

\noindent\textbf{Baselines.}
We organize baselines into two groups to evaluate different aspects of our method. Per-view segmentation methods (Mask2Former~\cite{mask2former22}, LSeg~\cite{lseg}) establish upper bounds on input-view segmentation, since they are applied directly to ground-truth images without any rendering. Joint NVS and scene understanding methods (LSM~\cite{lsm24}, SIU3R~\cite{SIU3R}) are the most direct comparisons, as they also target novel view panoptic segmentation from sparse unposed views.
Both construct explicit 3D Gaussian representations for label propagation, whereas our method propagates labels through the implicit correspondence in a frozen NVS model.
All learning-based baselines are re-trained on the same ScanNet split.

\noindent\textbf{Evaluation protocol.}
We evaluate segmentation on both input views and rendered novel views separately. Input view metrics measure source segmentation quality independent of the rendering pipeline. Novel view metrics evaluate the full system including panoptic propagation. 

\noindent\textbf{Metrics.}
For novel view synthesis, we report PSNR, SSIM, and LPIPS. For 
scene understanding, we report semantic mIoU and panoptic quality 
(PQ)~\cite{panopticseg19} defined as
\begin{equation}
  \mathrm{PQ} = \underbrace{\frac{\sum_{(p,g) \in \mathrm{TP}} 
  \mathrm{IoU}(p,g)}{|\mathrm{TP}|}}_{\mathrm{SQ}} \times 
  \underbrace{\frac{|\mathrm{TP}|}{|\mathrm{TP}| + 
  \frac{1}{2}|\mathrm{FP}| + 
  \frac{1}{2}|\mathrm{FN}|}}_{\mathrm{RQ}},
\end{equation}
where TP, FP, and FN denote matched, unmatched predicted, and  unmatched ground-truth segments, respectively. SQ measures boundary precision of matched segments, and RQ measures instance-level recognition. We report metrics on input views and on novel target views.

\begin{table*}[t]
\centering
\caption{\textbf{Comparisons of RGB reconstruction and panoptic segmentation on ScanNet~\cite{scannet}.} Input-view segmentation reflects the quality of the shared query decoder, and novel-view segmentation measures the combined effect of rendering and panoptic propagation.}
\label{tab:scannet}
% \resizebox{\textwidth}{!}{%
\begin{tabular}{l|ccc|cc|cc}
\toprule
 & \multicolumn{3}{c|}{Novel View Synthesis} & \multicolumn{2}{c|}{\shortstack{Input Views\\(2D-only)}} & \multicolumn{2}{c}{\shortstack{Novel Views\\(3D-aware)}} \\
\cmidrule(lr){2-4} \cmidrule(lr){5-6} \cmidrule(lr){7-8}
 & PSNR$\uparrow$ & SSIM$\uparrow$ & LPIPS$\downarrow$ & mIoU$\uparrow$ & PQ$\uparrow$ & mIoU$\uparrow$ & PQ$\uparrow$ \\

\midrule
\multicolumn{8}{l}{\textit{2D segmentation}} \\
Mask2Former~\cite{mask2former22} & - & - & - & \textbf{0.6186} & 0.5925 & - & - \\
LSeg~\cite{lseg} & - & - & - & 0.3976 & - & - & - \\
\midrule
\multicolumn{8}{l}{\textit{Joint NVS and scene understanding}} \\
LSM~\cite{lsm24} & 20.96 & 0.7245 & 0.3176 & 0.2810 & - & 0.2707 & - \\
SIU3R~\cite{SIU3R} & 25.88 & \uline{0.8220} & \uline{0.1831} & \uline{0.5899} & \textbf{0.6565} & \uline{0.5894} & \textbf{0.6565} \\
\textbf{Ours} & \textbf{33.56} & \textbf{0.9109} & \textbf{0.1149} & \textbf{0.6186} & \uline{0.5949} & \textbf{0.5949} & \uline{0.6092} \\

\bottomrule
\end{tabular}%
\end{table*}

\subsection{Main Results: Novel View Reconstruction and Segmentation}\label{sec:main_results}

Table~\ref{tab:scannet} shows the evaluation results of our method against baselines on ScanNet~\cite{scannet}. The results show that panoptic propagation preserves rendering quality while achieving competitive segmentation with a frozen novel view synthesis model.

\noindent\textbf{Panoptic propagation preserves rendering quality.}
Our method achieves 33.56~dB PSNR, the highest among all baselines. The segmentation path does not update the rendering weights, so RGB synthesis uses the same model as the original NVS path. The baselines SIU3R~\cite{SIU3R} and LSM~\cite{lsm24} couple reconstruction with understanding in a shared 3D Gaussian representation, reaching at most 25.88~dB. This separation preserves the rendering performance of the NVS backbone.

\noindent\textbf{Frozen propagation achieves competitive segmentation.}
On input views, our method matches the per-view Mask2Former~\cite{mask2former22} baseline in mIoU (0.6186) at the first row of Table~\ref{tab:scannet}, as both share the same query-based decoder architecture. On novel views, our method reaches 0.5949~mIoU, achieving better results than SIU3R~\cite{SIU3R} (0.5894) despite requiring no joint training.
At the last row, the mIoU drop from input views (0.6186) to novel views (0.5949) reflects propagation error. Binary channel encoding loses fine-grained boundary detail, and regions with low source-view overlap accumulate uncertainty.

SIU3R~\cite{SIU3R} shows better results than our method in terms of PQ (0.6565 vs.~0.6092). We attribute this gap to their multi-view mask aggregation mechanism, which rasterizes semantic attributes through 3D Gaussians and enforces cross-view instance consistency in 3D space. Our propagation operates in 2D token space without explicit geometric reasoning, making instance boundaries more susceptible to spatial ambiguity in the frozen attention.

\begin{figure*}[t]
    \centering
    \includegraphics[width=1.0\textwidth]{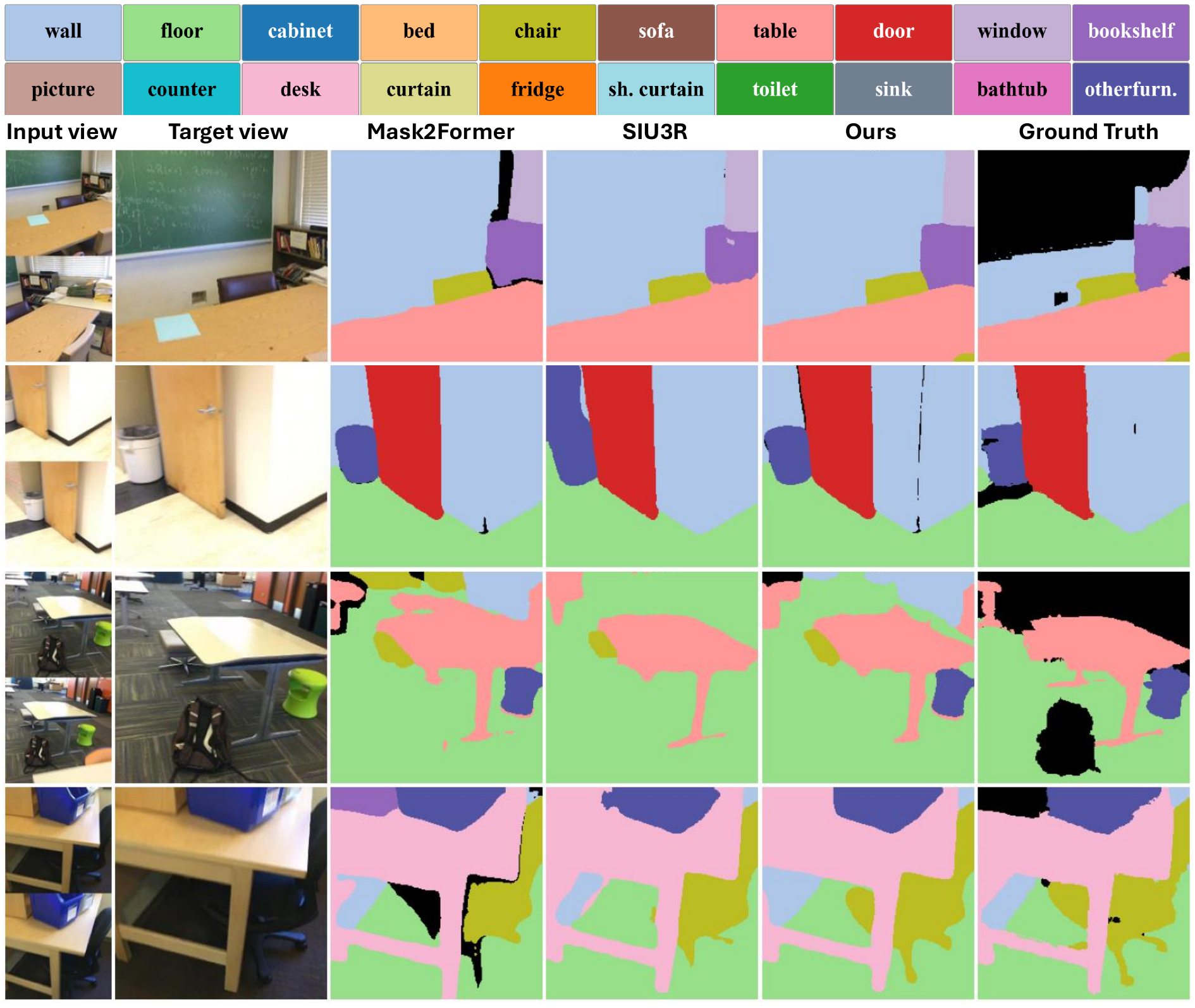}
    \caption{\textbf{Qualitative results of novel-view panoptic segmentation on ScanNet~\cite{scannet}.} From left to right: Input view, Target view, Panoptic segmentation results from Mask2Former~\cite{mask2former22}, SIU3R~\cite{SIU3R}, ours, and ground truth.} 
    \label{fig:qualtative}
\end{figure*}

%\pagebreak

\noindent\textbf{Qualitative comparison.}
Fig.~\ref{fig:qualtative} compares novel-view panoptic segmentation across methods. Our method propagates panoptic labels through the same frozen rendering model used for RGB synthesis, so the segmentation boundaries align with the rendered image appearance. SIU3R~\cite{SIU3R} produces segmentation through 3D Gaussian rasterization separately from RGB rendering, which can misalign the predicted labels with the rendered image. Mask2Former segments the rendered image directly, so any rendering imperfection propagates into the segmentation.

\begin{table}[t]
\centering
\caption{\textbf{Evaluation under different overlap ranges on ScanNet~\cite{scannet}.} Under low overlap, the PQ gap between SIU3R and ours narrows from 0.0473 to 0.0011, indicating that attention-based propagation degrades more gracefully than geometry-dependent rasterization. All the results use the same model reported in Table~\ref{tab:scannet}.}
\label{tab:low_overlap}
\setlength{\tabcolsep}{4pt}
\small
\begin{tabular}{@{}ll ccccc@{}}
\toprule
Overlap & Method 
  & PSNR$\uparrow$ & SSIM$\uparrow$ & LPIPS$\downarrow$ 
  & mIoU$\uparrow$ & PQ$\uparrow$ \\
\midrule
\multirow{2}{*}{High (0.3--0.8)}
  & SIU3R~\cite{SIU3R} 
  & 25.88 & 0.8220 & 0.1831 & 0.5894 & \textbf{0.6565} \\
  & Ours 
  & \textbf{33.56} & \textbf{0.9109} & \textbf{0.1149} 
  & \textbf{0.5949} & 0.6092 \\
\cmidrule(l){2-7}
  & \multicolumn{5}{r}{\footnotesize\textit{PQ gap}} 
  & \footnotesize 0.0473 \\
\midrule
\multirow{2}{*}{Low (0.01--0.3)}
  & SIU3R~\cite{SIU3R} 
  & 21.33 & 0.7147 & 0.2934 & 0.5458 & \textbf{0.5737} \\
  & Ours 
  & \textbf{26.47} & \textbf{0.7711} & \textbf{0.2544} 
  & \textbf{0.5477} & 0.5726 \\
\cmidrule(l){2-7}
  & \multicolumn{5}{r}{\footnotesize\textit{PQ gap}} 
  & \footnotesize 0.0011 \\
\bottomrule
\end{tabular}
\end{table}

\begin{table}[t]
\centering

\caption{\textbf{Results of design alternatives.} We evaluate several strategies in Fig.~\ref{fig:design_alternatives}. 
Rows 1 and 2 correspond to the architecture in Fig.~\ref{fig:design_alternatives}(a), where Row 1 does not freeze the parameters of the unposed NVS model but trainable. 
Row 3 corresponds to the design in Fig.~\ref{fig:design_alternatives}(b), while Row 4 corresponds to our proposed model in Fig.~\ref{fig:design_alternatives}(c). The column 2 (NVS) indicates whether the model parameters of the unposed NVS model were released for training (learnable) or kept frozen. }
%\emph{Mask2Former with Rendered Image} applies the frozen Mask2Former to our NVS output to isolate the effect of panoptic propagation.

\label{tab:design_ablations}

\setlength{\tabcolsep}{2pt}
\small
\begin{tabular}{@{}lcc ccccc@{}}
\toprule
Strategy & Fig. & NVS & PSNR$\uparrow$ & SSIM$\uparrow$ 
  & LPIPS$\downarrow$ & mIoU$\uparrow$ & PQ$\uparrow$ \\
\midrule
Joint decoder features & Fig.~\ref{fig:design_alternatives}(a) & Release 
  & 24.76 & 0.7502 & 0.2774 & 0.4146 & 0.4716 \\
Joint decoder features & Fig.~\ref{fig:design_alternatives}(a) & Freeze 
  & \textbf{33.56} & \textbf{0.9109} & \textbf{0.1149} 
  & 0.2008 & 0.1816 \\
Segment rendered image & Fig.~\ref{fig:design_alternatives}(b) & Freeze 
  & \textbf{33.56} & \textbf{0.9109} & \textbf{0.1149} 
  & \textbf{0.6239} & 0.5740 \\
\midrule
Ours (Propagation) & Fig.~\ref{fig:design_alternatives}(c) & Freeze  
  & \textbf{33.56} & \textbf{0.9109} & \textbf{0.1149} 
  & 0.5949 & \textbf{0.6092} \\
\bottomrule
\end{tabular}
\end{table}

\subsection{Emerging Properties at Low-Overlap Scenarios}\label{sec:low_overlap}

Table~\ref{tab:low_overlap} evaluates robustness under sparse geometric overlap, using view pairs with depth-based IoU in $[0.01, 0.3]$ following SIU3R~\cite{SIU3R}.
This setting stresses label propagation because large target regions are weakly observed or dis-occluded from the source views, making explicit geometry estimation less reliable.
Our method maintains a 5.14~dB PSNR advantage and slightly higher mIoU under this setting.
More importantly, the PQ gap to SIU3R shrinks from 0.0473 under standard overlap to 0.0011 under low overlap.
This suggests that the frozen NVS transformer retains a learned extrapolation capability. Even when explicit reconstruction becomes unreliable under sparse overlap, the token-level correspondence still transfers useful panoptic structure.

\begin{figure*}[t]
    \centering
    \includegraphics[width=1.0\textwidth]{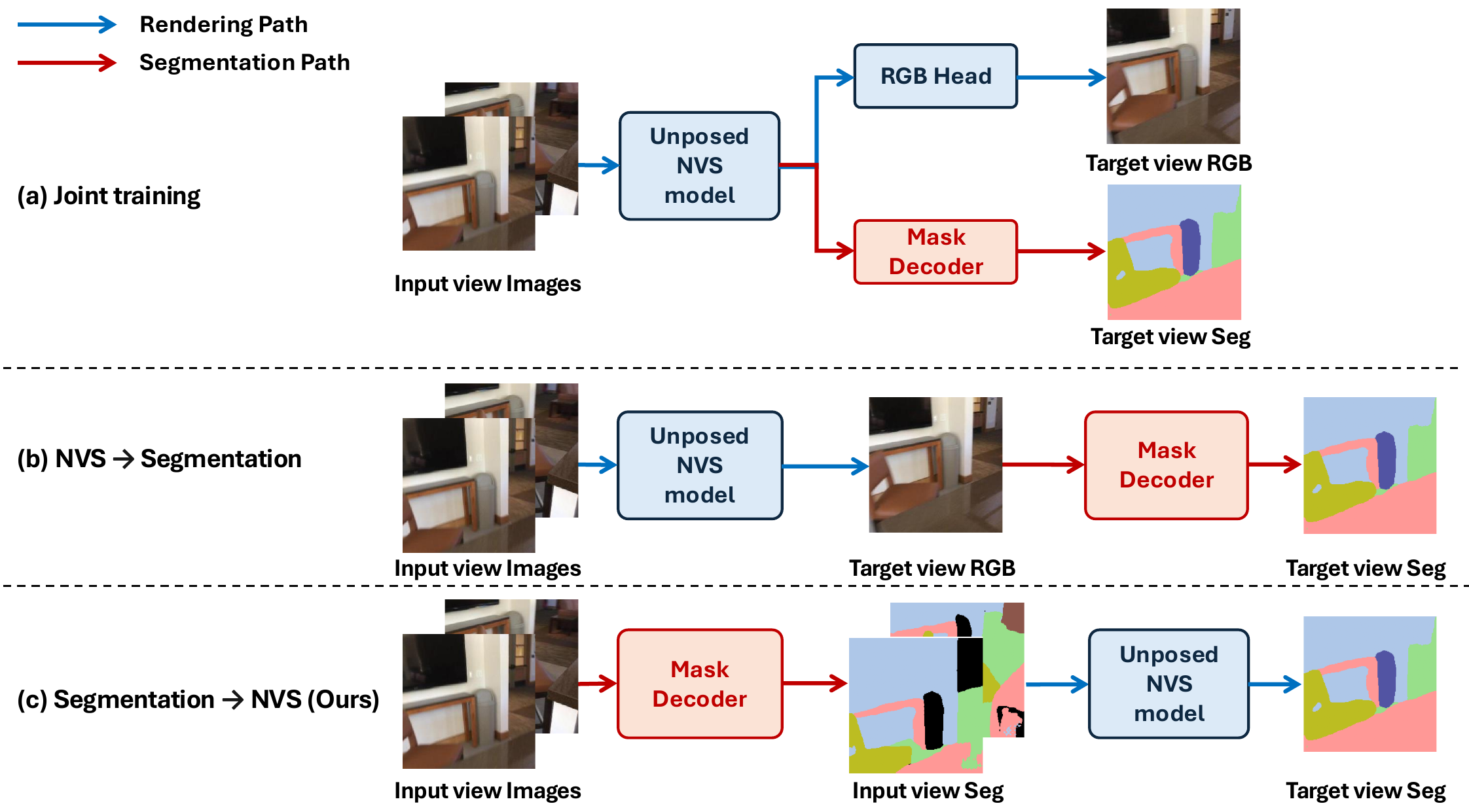}
    \caption{\textbf{Design alternatives.} (a)~Joint training attaches a segmentation head to intermediate features of the NVS rendering decoder. (b)~NVS$\rightarrow$Segmentation applies a per-view segmenter, Mask2Former~\cite{mask2former22}, to the rendered novel-view image. (c)~Our method first segments the input views, encodes the panoptic maps into binary channel representations, and propagates them through the frozen NVS model to produce target-view segmentation. Quantitative comparisons are reported in Table~\ref{tab:design_ablations}.}
    \label{fig:design_alternatives}
\end{figure*}

\subsection{Design Alternatives for Novel-View Panoptic Segmentation}\label{sec:design_alternatives}

Table~\ref{tab:design_ablations} compares three strategies for integrating panoptic segmentation with the NVS model~\cite{less3depend}, validating the design of our decoupled propagation approach. The three strategies, illustrated in Fig.~\ref{fig:design_alternatives}, differ in how segmentation is integrated with the rendering pipeline.

The first strategy (Fig.~\ref{fig:design_alternatives}(a)) attaches a segmentation head to intermediate features of the frozen rendering decoder. As shown in the second row of Table~\ref{tab:design_ablations}, these features are optimized for appearance reconstruction rather than semantic discrimination, causing segmentation performance to collapse to 0.2008~mIoU and 0.1816~PQ.
In the first row of Table~\ref{tab:design_ablations}, allowing the NVS model to be jointly optimized with the segmentation head partially recovers segmentation accuracy, but introduces interference between photometric and segmentation objectives. This joint training degrades both rendering quality (24.76~dB) and segmentation performance (0.4146~mIoU, 0.4716~PQ).

The second strategy (Fig.~\ref{fig:design_alternatives}(b)) applies a per-view segmenter~\cite{mask2former22} to each rendered novel-view image independently. While this avoids modifying the NVS model, the third row of Table~\ref{tab:design_ablations} shows that PQ drops to 0.5740 due to inconsistent instance IDs across views, indicating that high rendering quality alone does not ensure cross-view panoptic consistency.

Finally, our method (Fig.~\ref{fig:design_alternatives}(c)) propagates panoptic logits decoded from the source views through the frozen NVS model. As shown in the fourth row of Table~\ref{tab:design_ablations}, this decoupled design preserves the full rendering quality (33.56~dB) while achieving the best panoptic quality (0.6092 PQ), outperforming joint-training alternatives without compromising rendering fidelity.

\begin{table}[t]
\centering
\caption{{\textbf{Modular composition and cross-dataset transfer.} The Segmenter column lists the source-view segmenter and its encoder. SIU3R and our \textit{M2F} variant share a Mask2Former~\cite{mask2former22} head but differ in encoder, CroCo~\cite{WeinzaepfelNIPS22CroCoSelfSupervisedPreTraining4CrossViewCompletion} for SIU3R and DINOv2~\cite{dinov2} for ours. \textit{PanSt3R}~\cite{panst3r25} is an off-the-shelf predictor that adds no segmentation-loss training. ``Pose'' indicates whether the target view uses the ground-truth pose (GT) or the estimated latent Pl\"ucker embedding (latent). We evaluate Replica~\cite{StraubX19Replica} zero-shot, without any Replica fine-tuning.}}
\label{tab:modular_replica}
\setlength{\tabcolsep}{5pt}
\small
\resizebox{\textwidth}{!}{%
\begin{tabular}{@{}lll cccc cccc@{}}
\toprule
\multirow{2}{*}{Method} & \multirow{2}{*}{Segmenter} & \multirow{2}{*}{Pose}
  & \multicolumn{4}{c}{ScanNet~\cite{scannet}}
  & \multicolumn{4}{c}{Replica~\cite{StraubX19Replica} (zero-shot)} \\
\cmidrule(lr){4-7} \cmidrule(lr){8-11}
 & & & PSNR$\uparrow$ & SSIM$\uparrow$ & PQ$\uparrow$ & mIoU$\uparrow$
     & PSNR$\uparrow$ & SSIM$\uparrow$ & PQ$\uparrow$ & mIoU$\uparrow$ \\
\midrule
SIU3R~\cite{SIU3R} & M2F (CroCo) & GT
  & 25.88 & 0.822 & \textbf{0.657} & 0.589
  & 14.05 & 0.527 & 0.186 & 0.171 \\
\midrule
\multirow{3}{*}{Ours} & M2F (DINOv2) & latent
  & \textbf{33.56} & \textbf{0.911} & 0.609 & \textbf{0.595}
  & -- & -- & -- & -- \\
\cmidrule(lr){2-11}
 & \multirow{2}{*}{PanSt3R~\cite{panst3r25}} & GT
  & 25.02 & 0.782 & 0.510 & 0.366
  & 21.89 & 0.751 & 0.246 & 0.256 \\
 & & latent
  & 28.28 & 0.850 & 0.580 & 0.593
  & \textbf{23.52} & \textbf{0.786} & \textbf{0.394} & \textbf{0.454} \\
\bottomrule
\end{tabular}%
}
\end{table}

\begin{figure}[t]
    \centering
    \includegraphics[width=1.0\textwidth]{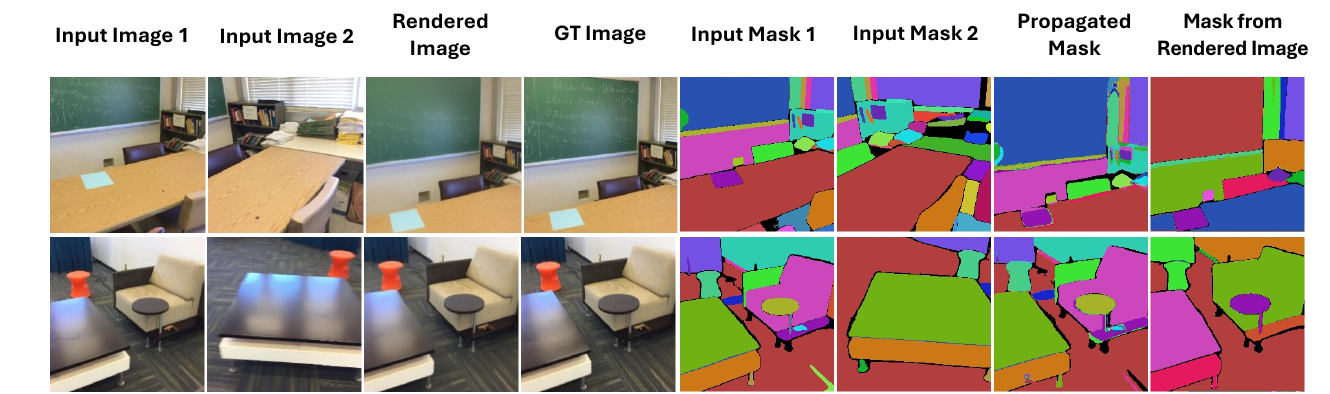}
    \caption{
        \textbf{Replacing the input view segmentation model.}
        The novel view synthesis model propagates SAM2~\cite{sam2} predictions (Input Mask 1 and 2) to the novel viewpoint (Propagated Mask) while preserving instance identity across views.
    }
    \label{fig:sam2}
\end{figure}

\subsection{Modular Composition and Cross-Dataset Transfer}\label{sec:modular_replica}

Our decoupled design lets us replace the source-view segmenter with an off-the-shelf predictor without retraining the rest of the system. 
We choose PanSt3R~\cite{panst3r25} because it produces source-view panoptic labels that transfer across datasets, which lets us test propagation under cross-dataset evaluation.
We replace our shared query decoder with it to supply source-view labels, testing whether the frozen propagation path generalizes beyond our trained segmentation module, and report the results in Table~\ref{tab:modular_replica}.
On ScanNet, \textit{Ours(PanSt3R)} stays competitive without any ScanNet fine-tuning of the segmenter, and \textit{Ours(M2F)} keeps the rendering advantage over SIU3R~\cite{SIU3R}. 
On Replica, \textit{Ours(PanSt3R)} outperforms SIU3R~\cite{SIU3R} on every RGB and segmentation metric (23.52 vs.\ 14.05~dB PSNR, 0.394 vs.\ 0.186 PQ, 0.454 vs.\ 0.171 mIoU). 
The higher ScanNet PQ of SIU3R~\cite{SIU3R} thus partly reflects in-domain specialization, while reconstruction-free propagation transfers better across datasets. 
We therefore frame our contribution as reconstruction-free label propagation with high rendering quality and transferable segmentation, not as overall PQ superiority.
The latent-query variant matches or exceeds the GT-pose variant on both datasets. The estimated latent Pl\"ucker embedding therefore carries enough view information for instance-level propagation.

The same modularity holds for a class-agnostic segmenter. As the source-view segmenter, we replace Mask2Former~\cite{mask2former22} with SAM2~\cite{sam2} and propagate its masks through the same frozen NVS model. Fig.~\ref{fig:sam2} shows that the propagated masks preserve instance identity across views with comparable visual quality. These results show that the propagation stage can take labels from different source-view segmenters without retraining.

\section{Conclusion}
This work demonstrates that a view synthesis model, trained for novel view RGB image reconstruction, propagates panoptic labels to novel viewpoints with quality comparable to methods that build explicit 3D representations with dedicated segmentation objectives. 
The only learned segmentation component is a shared query decoder that produces source-view labels, and the frozen rendering transformer performs the cross-view propagation.
The implicit correspondence inside the transformer thus generalizes beyond appearance and already encodes enough geometric structure to transfer arbitrary per-pixel signals. 
This result suggests that large view synthesis models can serve as an alternative propagation mechanism to explicit 3D reconstruction for novel-view panoptic labels.
As long as the transferred signal stays valid across viewpoints, large view synthesis models may also be applicable to other view-consistent dense prediction signals.

\noindent\textbf{Limitation.}
Our pipeline requires a full forward pass of the view synthesis model for every target view, which becomes a bottleneck when only segmentation at a known viewpoint is needed. The propagation also assumes that the transferred signal stays valid across viewpoints. This holds for semantic classes and instance identities. View-dependent quantities such as depth do not satisfy this assumption, and extending the propagation to them would require a camera-aware transformation within the propagation path.

\smallskip
\noindent\textbf{Acknowledgement.}
This work was supported by IITP grant (RS-2021-II211343: AI Graduate School Program at Seoul National Univ. (5\%), RS-2025-25442338: AI star Fellowship Support Program at Seoul National Univ. (20\%), and RS-2026-25517417: Development of Compression, Reconstruction, and Rendering Technologies for Free-viewpoint Media. (75\%)).
POSCO DX Company, Ltd. provided generous support for this research.

\clearpage
\hypersetup{pageanchor=false}
\setcounter{page}{1}

\setcounter{section}{0}
\setcounter{table}{0}
\setcounter{figure}{0}

\title{\textit{Supplementary Material} for Extending a Large View Synthesis Model for Multi-view Panoptic Segmentation}

\author{}
\institute{}

\titlerunning{Extending LVSM for Panoptic Seg}
\authorrunning{K. Ryu et al.}
\maketitle

\appendix
\setcounter{equation}{0}
\renewcommand{\theHsection}{supplement.\arabic{section}}
\renewcommand{\theHsubsection}{supplement.\arabic{section}.\arabic{subsection}}
\renewcommand{\theHfigure}{supplement.figure.\arabic{figure}}
\renewcommand{\theHtable}{supplement.table.\arabic{table}}
\renewcommand{\theHequation}{supplement.equation.\arabic{equation}}
In this supplementary material, we provide additional experiments and analyses that complement the main paper.
Specifically, we present the following additional results.

\begin{enumerate}
    \item Experimental setup details including depth-based IoU for view pair sampling and architecture configurations (Sec.~\ref{sec:experimental_setup}).
    \item Multi-pass decoding for scenes with more than eight instances (Sec.~\ref{sec:multipass}).
    \item Ablation on instance encoding strategies and the effect of the uncertainty threshold $\tau$(Sec.~\ref{sec:encoding}).
    \item Ablation on NVS backbone choice to validate our pipeline works with different NVS backbones (Sec.~\ref{sec:nvs_backbone}).
    \item Generalization to more input views ($V_i{=}3$) without architectural change (Sec.~\ref{sec:nview}).
    \item Details and additional examples of gradient saliency analysis (Sec.~\ref{sec:analysis_supple}).
    \item Additional qualitative results, including comparison under low overlap, SAM2 mask propagation, outdoor-scene generalization, and supplementary video (Sec.~\ref{sec:qualitative_results}).
\end{enumerate}

\section{Details for the Experimental setup}\label{sec:experimental_setup}

In this section we provide detailed experiment setup, specifically for the pair sampling.
Following SIU3R~\cite{SIU3R}, we train and evaluate on ScanNet~\cite{scannet} with image pairs sampled by depth-based IoU. 
For each view pair $(i, j)$, we unproject valid pixels into 3D using their depth maps and reproject onto the other view. 
The depth-based IoU is defined as follows,
\begin{equation}
  \text{IoU}_{\text{depth}}(i, j) 
  = \frac{|\mathcal{V}_{i \to j}| + |\mathcal{V}_{j \to i}|}
         {|\mathcal{P}_i| + |\mathcal{P}_j|},
  \label{eq:depth_iou}
\end{equation}
where $\mathcal{V}_{i \to j}$ is the set of pixels in view $i$ visible in view $j$, and $\mathcal{P}_i$ is the set of all valid pixels in view $i$.
We sample pairs with $\text{IoU}_{\text{depth}} \in [0.3, 0.8]$ for both training and evaluation. In low-overlapped scenario, we sample pairs with $\text{IoU}_{\text{depth}} \in [0.01, 0.3]$.
We also provide the full architecture and training details in Tab.~\ref{tab:implementation} for NVS model and input view segmentation model. 

\begin{table}[t]
\centering
\caption{Architecture and training details.}
\label{tab:implementation}
\begin{tabular}{ll}
\toprule
\multicolumn{2}{l}{\textbf{NVS model}} \\
\midrule
Backbone & DINOv2(ViT-B/14) \\
Encoder / Pose Estimator / Decoder & 6 / 4 / 14 layers \\
Latent vectors dimension & 1024 \\
\midrule
\multicolumn{2}{l}{\textbf{Input view segmenter}} \\
\midrule
Seg encoder & DINOv2(ViT-B/14) \\
Seg adapter & ViT-Adapter (12 blocks) \\
Seg head & VideoMask2Former (100 queries, 20 classes) \\
\midrule
\multicolumn{2}{l}{\textbf{Training}} \\
\midrule
Optimizer & AdamW  \\
Base / Min LR & 1e-4 / 1e-5 \\
Scheduler & Cosine annealing, 2 epoch warmup \\
Weight decay & 0.05 \\
$\beta_1$ / $\beta_2$ & 0.9 / 0.95 \\
Gradient clipping & 1.0 \\
Precision & bf16 (AMP) \\
\bottomrule
\end{tabular}
\end{table}

\noindent\textbf{Training objective.}
We fine-tune the NVS model with the photometric loss $\mathcal{L}_{\mathrm{MSE}} + \lambda \mathcal{L}_{\mathrm{perceptual}}$ and set $\lambda = 0.5$ following Less3Depend~\cite{less3depend}. We train the shared query decoder with the standard Mask2Former~\cite{mask2former22} objective on the input-view labels,
\begin{equation}
  \mathcal{L}_{\mathrm{seg}} = \lambda_{\mathrm{ce}}\,\mathcal{L}_{\mathrm{ce}}
  + \lambda_{\mathrm{dice}}\,\mathcal{L}_{\mathrm{dice}}
  + \lambda_{\mathrm{cls}}\,\mathcal{L}_{\mathrm{cls}},
\end{equation}
where $\mathcal{L}_{\mathrm{ce}}$ and $\mathcal{L}_{\mathrm{dice}}$ are the cross-entropy and dice losses for mask prediction and $\mathcal{L}_{\mathrm{cls}}$ is the classification loss, all computed over Hungarian-matched pairs. We set $\lambda_{\mathrm{ce}} = 5.0$, $\lambda_{\mathrm{dice}} = 5.0$, $\lambda_{\mathrm{cls}} = 2.0$ following Mask2Former~\cite{mask2former22}.

\section{Handling More Than Eight Instances}\label{sec:multipass}
The 3-bit code represents at most eight active instance IDs in a single propagation pass, but this is a per-pass capacity rather than a hard scene-level limit. 
For a scene with $K > 8$ instances, we partition the IDs into $M = \lceil K / 8 \rceil$ groups, encode each group with a local 3-bit lookup table, and propagate each group through the same frozen view synthesis model. 
We then map the decoded local IDs back to global IDs through the lookup table and merge them into a single panoptic map. 
This multi-pass strategy keeps the wide inter-codeword margin that makes binary encoding robust. 
It increases inference time linearly with $M$ and needs no retraining or architectural change. Scenes with more than eight instances are rare in our evaluation, so a single pass suffices for most view pairs.

\section{Ablation on Instance Encoding Strategies}\label{sec:encoding}

We compare several strategies for encoding instance IDs into 3-channel images, including binary encoding, color grid encoding, and HSV color mapping. 
Let $G$ denote the number of grid intervals per channel, so that each channel takes $(G{+}1)$ discrete levels and up to $(G{+}1)^3$ instances can be represented. 
We refer readers to Sec.~3.4 in main paper for ablations justifying our encoding design.

\begin{table}[t]
\centering
\caption{Comparison of instance encoding strategies on ScanNet, using \textit{Ours~(M2F)}. 
NVS quality is identical across all methods (shared model). 
Binary ($G{=}1$, $\tau{=}0.2$) achieves the best segmentation.}
\label{tab:color_grid}
\setlength{\tabcolsep}{5pt}
\begin{tabular}{l@{\hskip 6pt}c@{\hskip 6pt}c|cc|cc}
\toprule
\multicolumn{3}{l|}{\textit{Shared NVS quality}} & \multicolumn{4}{c}{PSNR 33.56 / SSIM 0.9109 / LPIPS 0.1149} \\
\midrule
\multirow{2}{*}{Method} & \multirow{2}{*}{$G$} & \multirow{2}{*}{$\tau$} & \multicolumn{2}{c|}{Input Views} & \multicolumn{2}{c}{Novel Views} \\
& & & mIoU & PQ & mIoU & PQ \\
\midrule
\textbf{Binary} & \textbf{1} & \textbf{0.2} & \textbf{0.6186} & \textbf{0.5949} & \underline{0.5949} & \textbf{0.6092} \\
Binary & 1 & 0.0 & \underline{0.6104} & \underline{0.5567} & \textbf{0.6045} & \underline{0.5449} \\
Grid & 3 & 0.0 & 0.6037 & 0.5363 & 0.5977 & 0.5270 \\
Grid & 4 & 0.0 & 0.6004 & 0.5262 & 0.5939 & 0.5152 \\
Grid & 5 & 0.0 & 0.5968 & 0.5159 & 0.5898 & 0.5023 \\
HSV & - & - & 0.5740 & 0.4603 & 0.5642 & 0.4436 \\
\bottomrule
\end{tabular}
\end{table}

\noindent\textbf{Binary encoding.}
Our default encoding uses $G{=}1$, mapping each channel to two levels 
(0.1 and 0.3). 
A threshold $\tau{=}0.2$ rejects rendered values in 
the ambiguous range $[0.3, 0.7]$. 
As shown in Tab.~\ref{tab:color_grid}, 
this simple rejection improves both mIoU and PQ over $\tau{=}0.0$.

\noindent\textbf{Color grid encoding.}
Subdividing each channel into $G{>}1$ levels increases capacity 
(up to $(G{+}1)^3$ instances), but reduces the color margin between adjacent levels. 
The NVS model blends colors at object boundaries, and with a finer grid these blended values easily cross into neighboring grid cells. 
This fragments contiguous instances into disconnected segments, a failure we call \textit{hollow instances}. 
This differs from the phantom instances in Sec.~3.4, which arise when blended values alias into a valid but incorrect codeword. 
Tab.~\ref{tab:color_grid} confirms that PQ drops consistently as $G$ increases from 3 to 5. 

\noindent\textbf{HSV encoding.}
Encoding instance IDs via uniformly spaced hues is a common visualization choice, but the circular and nonlinear nature of the hue channel makes it fragile. 
A small color shift after rendering can map to a completely different instance ID, making hollow instances far more frequent. 
HSV yields the lowest PQ among all methods (Tab.~\ref{tab:color_grid}).
We choose binary encoding because with $G{=}1$ the two levels are maximally separated, and ambiguous values cluster near 0.5, making them easy to reject with $\tau$.

\subsection{Effect of the Uncertainty Threshold}\label{sec:tau_sweep}
The threshold $\tau$ controls how aggressively we reject ambiguous boundary pixels, where any channel satisfying $|\hat{b}_{:,:,k}(p) - 0.5| \leq \tau$ is left unlabeled. 
We sweep $\tau$ using \textit{Ours~(PanSt3R, latent)} on ScanNet and report metrics both including ($i$) and excluding ($e$) uncertain pixels from evaluation in Tab.~\ref{tab:tau_sweep}. 
At our default $\tau{=}0.20$, only $3.4\%$ of pixels are marked uncertain, and across the practical range $\tau\in[0.10, 0.30]$ the uncertain fraction stays below $6\%$ and the conclusion remains stable. 
PQ and mIoU respond to uncertain-pixel rejection in opposite ways. 
Increasing $\tau$ improves PQ by removing noisy boundary segments, but lowers mIoU once a large fraction of pixels is left unlabeled. 
Our default $\tau{=}0.20$ balances this asymmetry, and the include/exclude policy (PQ$_i$ vs.\ PQ$_e$) does not change the ranking.

\begin{table}[t]
\centering
\caption{Effect of the uncertainty threshold $\tau$ on ScanNet, using \textit{Ours~(PanSt3R, latent)}. ``uncert.\%'' is the fraction of pixels left unlabeled. Subscripts $i$ and $e$ include and exclude uncertain pixels from evaluation. $\dagger$ marks our default.}
\label{tab:tau_sweep}
\setlength{\tabcolsep}{8pt}
\small
\begin{tabular}{@{}cccccc@{}}
\toprule
$\tau$ & uncert.\% & PQ$_{i}$ & PQ$_{e}$ & mIoU$_{i}$ & mIoU$_{e}$ \\
\midrule
0.10           & 1.6  & 0.560 & 0.559 & 0.599 & 0.609 \\
0.20$^{\dagger}$ & 3.4  & 0.580 & 0.578 & 0.593 & 0.612 \\
0.30           & 6.0  & 0.606 & 0.599 & 0.577 & 0.612 \\
0.40           & 21.2 & 0.657 & 0.626 & 0.419 & 0.526 \\
0.45           & 41.2 & 0.727 & 0.657 & 0.189 & 0.338 \\
\bottomrule
\end{tabular}
\end{table}

\section{Ablation on NVS Backbone}\label{sec:nvs_backbone}

We also validate whether our pipeline works with a different NVS backbone by swapping Less3Depend~\cite{less3depend} with RayZer~\cite{rayzer}. 
Both handle sparse, unposed images, so they are drop-in replacements. 
Tab.~\ref{tab:nvs_ablation} shows that Less3Depend~\cite{less3depend} achieves higher rendering quality on ScanNet, which directly leads to better segmentation propagation.

\begin{table}[t]
  \centering
  \caption{Ablation on NVS backbone on ScanNet~\cite{scannet}.}
  \label{tab:nvs_ablation}
  \begin{tabular}{l ccc cc}
    \toprule
    & \multicolumn{3}{c}{NVS Quality} & \multicolumn{2}{c}{Seg. Propagation} \\
    \cmidrule(lr){2-4} \cmidrule(lr){5-6}
    NVS Backbone & PSNR$\uparrow$ & SSIM$\uparrow$ & LPIPS$\downarrow$ & mIoU$\uparrow$ & PQ$\uparrow$ \\
    \midrule
    RayZer~\cite{rayzer} & 28.47 & 0.8578 & 0.2018 & 0.4916 & 0.5523 \\
    Less3Depend~\cite{less3depend} (Ours) & 33.56 & 0.9109 & 0.1149 & 0.5949 & 0.6092 \\
    \bottomrule
  \end{tabular}
\end{table}

\section{Generalization to More Input Views}\label{sec:nview}

Since the underlying view synthesis model renders from a variable number of source views, our pipeline extends to more than two input views without any architectural change. 
We additionally evaluate a $V_i{=}3$ setting using the corresponding 3-view evaluation protocol on ScanNet~\cite{scannet}. 
As reported in Tab.~\ref{tab:nview}, the pipeline operates across both view counts, showing that the same propagation procedure runs with both two and three input views. 
Because the 3-view protocol differs from the default 2-view split (\eg, in sampled pairs and overlap distribution), the two rows reflect different evaluation settings rather than a controlled view-count ablation, and the score change should be read as compatibility evidence rather than a strict scaling trend.

\begin{table}[t]
  \centering
  \caption{Generalization to more input views on ScanNet~\cite{scannet}. The pipeline supports $V_i{=}3$ without architectural modification.}
  \label{tab:nview}
  \setlength{\tabcolsep}{8pt}
  \small
  \begin{tabular}{l cccc}
    \toprule
    Setting & mIoU$\uparrow$ & PQ$\uparrow$ & PSNR$\uparrow$ & SSIM$\uparrow$ \\
    \midrule
    $V_i{=}2$ (default) & 0.5949 & 0.6092 & 33.56 & 0.9109 \\
    $V_i{=}3$           & 0.5440 & 0.5711 & 32.43 & 0.8959 \\
    \bottomrule
  \end{tabular}
\end{table}

\section{Details of Gradient Saliency}\label{sec:analysis_supple}

In Sec.~3.2, we analyze the implicit correspondence of the NVS transformer via gradient saliency. 
We provide the full formulation and additional visualization examples in Fig.~\ref{fig:saliency_supple}.

\subsection{Formulation}

Let $\{I^\mathrm{s}_i\}_{i=1}^{N}$ denote $N$ input views. 
The DINOv2 tokenizer in the scene encoder maps each view to patch tokens $\mathbf{Z}^\mathrm{s} \in \mathbb{R}^{N \cdot N_p \times d}$, where $N_p = (H/p)^2$ is the number of patches per view and $d{=}768$. 

Given a target query patch at index $q$, the scene encoder $\mathcal{E}$ and render decoder $\mathcal{R}$ produce the output $\hat{\mathbf{y}}_q \in \mathbb{R}^{p^2 \cdot 3}$ as described in Sec.~3.1 of the main paper. We define the gradient saliency of source token $(v, i)$ as the average absolute gradient of the channel-summed output with respect to each feature dimension,
\begin{equation}
    S_{q}(v, i) = \frac{1}{d} \sum_{k=1}^{d} \left| 
    \frac{\partial f_q}{\partial \, z^{(v,i)}_{k}} \right|, 
    \quad f_q = \sum_{c} \hat{y}_{q,c},
    \label{eq:saliency}
\end{equation}
where $z^{(v,i)}_{k}$ is the $k$-th feature of the source token at patch $i$ in view $v$. 
Unlike attention rollout~\cite{abnar2020quantifying}, which only captures attention weight propagation, gradient saliency captures the full computational graph including value projections, MLPs, and residual connections. 

\subsection{Comparison in different modality}

We compute saliency maps for two input modalities through the same  NVS pipeline. 
Source images $\{I^\mathrm{s}_i\}_{i=1}^{N}$ are tokenized and processed through $\mathcal{E}$ and $\mathcal{R}$ for RGB saliency $S_q^\mathrm{rgb}$. 
For binary saliency $S_q^\mathrm{seg}$, binary encoded instance maps $\mathbf{b}^\mathrm{s}_i = \mathcal{B}(\mathbf{M}^\mathrm{s}_i)$ are tokenized and processed through the same $\mathcal{E}$ and $\mathcal{R}$ with the segmentation latent $\mathbf{z}_0^\mathrm{seg}$.

\subsection{Observation}
\begin{figure}[t]
    \centering
    \includegraphics[width=\textwidth]{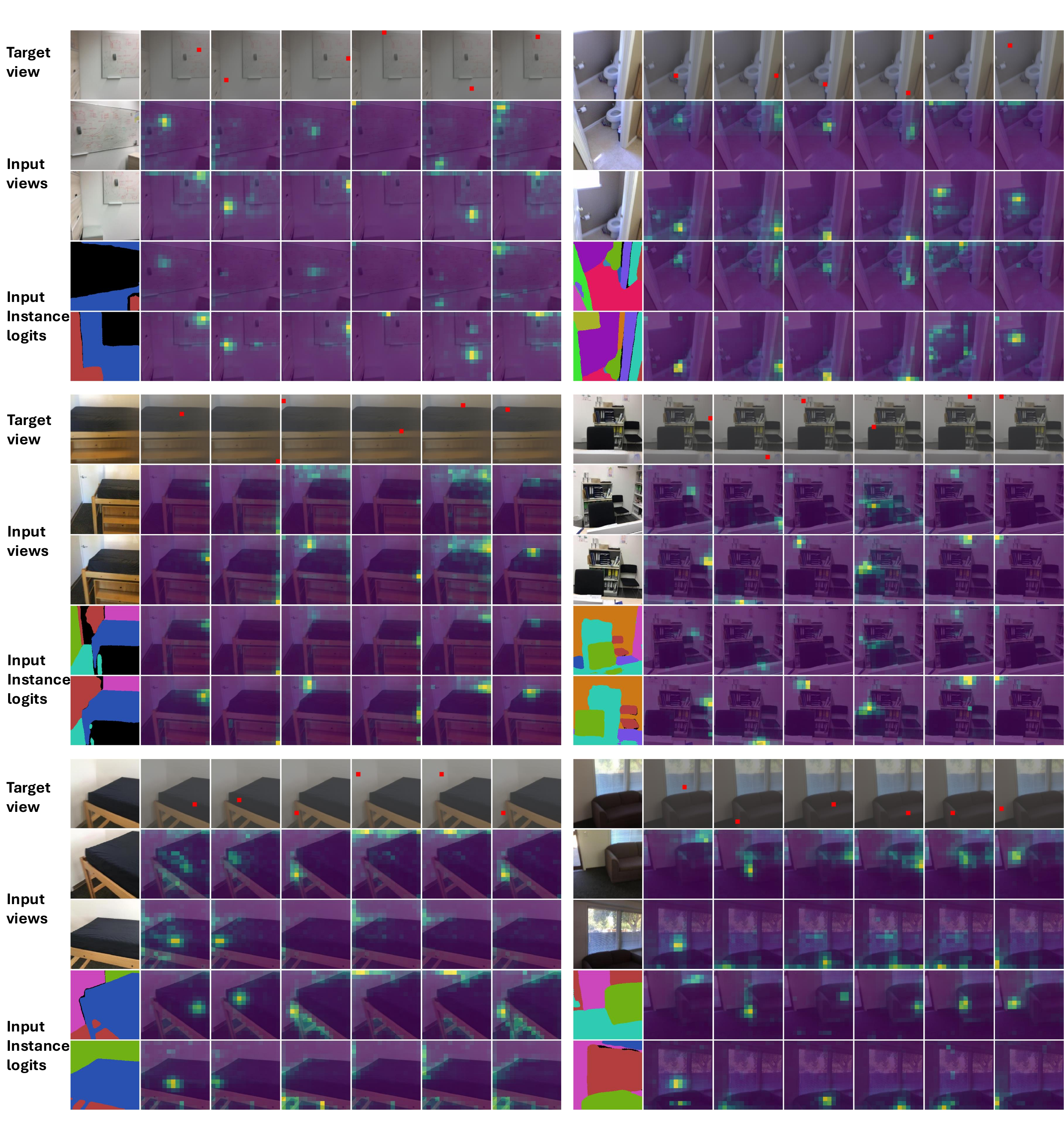}
    \vspace{-0.6cm}
    \caption{
        \textbf{Additional visualization of gradient saliency from target to input views.} 
        Row 1 shows the rendered target view, with each column indicating the location of a different query patch in red. For each query patch, Rows 2 and 3 and Rows 4 and 5 show the gradient saliency over the same two input views, using RGB inputs and binary instance encodings, respectively.
    }
    \label{fig:saliency_supple}
\end{figure}

As shown in Fig.~\ref{fig:saliency_supple}, the saliency concentrates on geometrically corresponding regions in the input views for a given target query $q$. 
As discussed in Sec.~3.2 of the main paper, the saliency maps show consistent patterns across RGB and binary instance inputs, indicating that the model resolves correspondence based on geometric pose rather than input content. 
This is the key property that enables panoptic propagation through the frozen NVS model without segmentation training. 
We also observe that the saliency differs between the two input views, as the model selectively attends more to whichever view better observes the queried target region.

\section{Additional Qualitative Results}\label{sec:qualitative_results}
In this section, we present qualitative comparisons with SIU3R under low overlap, SAM2 mask propagation results, generalization to outdoor scenes, and a supplementary video.

\begin{figure}[t]
    \centering
    \includegraphics[width=\textwidth]{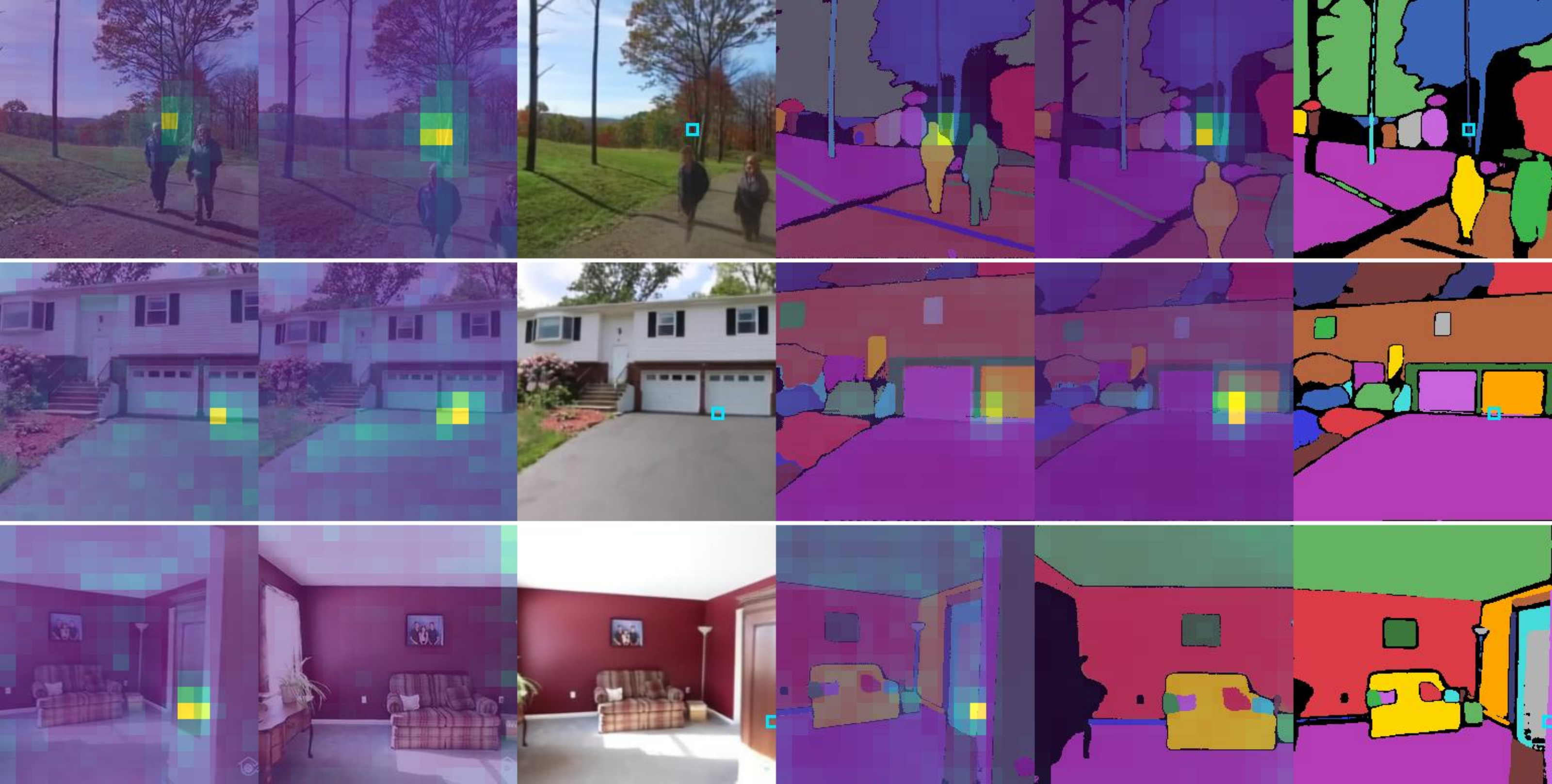}
    \caption{\textbf{Qualitative results on outdoor scenes from RealEstate10K~\cite{re10k}.} Our pipeline propagates source-view panoptic labels to novel views with consistent instance identities beyond indoor ScanNet scenes.}
    \label{fig:qualitative_re10k}
\end{figure}

\subsection{Generalization to Outdoor Scenes}
To probe whether the propagation mechanism is tied to the indoor geometry of ScanNet~\cite{scannet}, we provide qualitative results on outdoor scenes from RealEstate10K ~\cite{re10k} in Fig.~\ref{fig:qualitative_re10k}. 
The frozen view synthesis model propagates the source-view panoptic labels to novel viewpoints with consistent instance identities, despite the larger baselines and different scene statistics of outdoor captures. 
This indicates that the learned implicit correspondence, and thus our reconstruction-free propagation, is not specific to indoor scene types.

\subsection{Comparison under Low Overlap}
\begin{figure}[t]
    \centering
    \includegraphics[width=\textwidth]{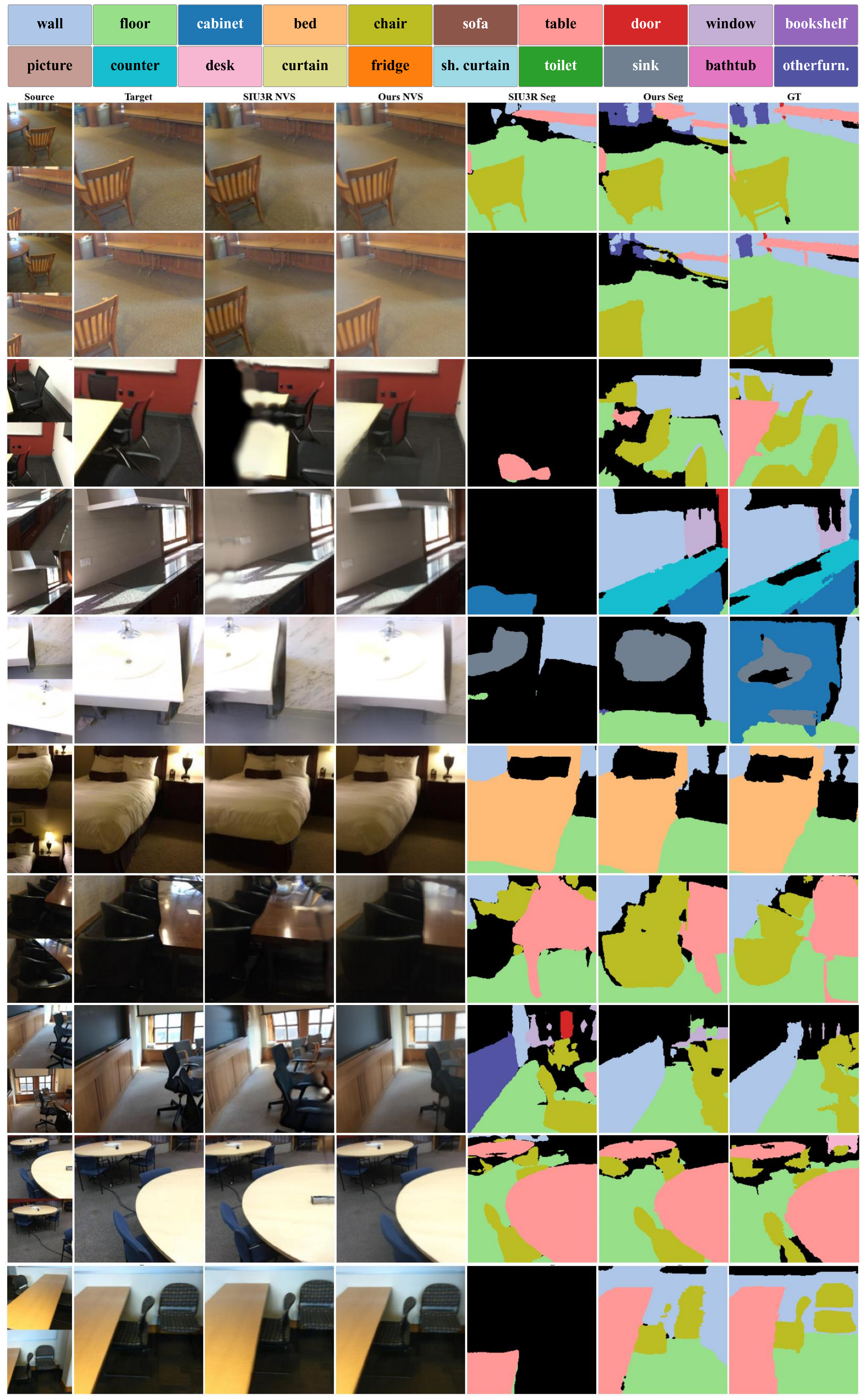}
    \caption{
        \textbf{Additional qualitative comparison with SIU3R~\cite{SIU3R} in low-overlapped scenario.} Our method maintains both rendering and segmentation quality while SIU3R suffers from pose estimation errors. 
    }
    \label{fig:qualitative_supple}
\end{figure}

Figures~\ref{fig:qualitative_supple} show qualitative comparisons between our method and SIU3R~\cite{SIU3R} on novel view panoptic segmentation in low overlapped scenarios. SIU3R~\cite{SIU3R} relies on MASt3R~\cite{mast3r} for 3D point cloud estimation and constructs 3D Gaussians for rendering and segmentation. However, the point clouds from MASt3R~\cite{mast3r} suffer from scale ambiguity, and the normalized poses used for Gaussian Splatting rendering do not accurately reflect the true camera geometry, directly degrading both rendering and segmentation quality. The panoptic postprocessing applies confidence thresholds to mask logits, and Gaussian-based methods produce noisier opacity from the accumulated pose errors, so many masks are discarded during this filtering step.

\begin{figure}[t]
    \centering
    \includegraphics[width=\textwidth]{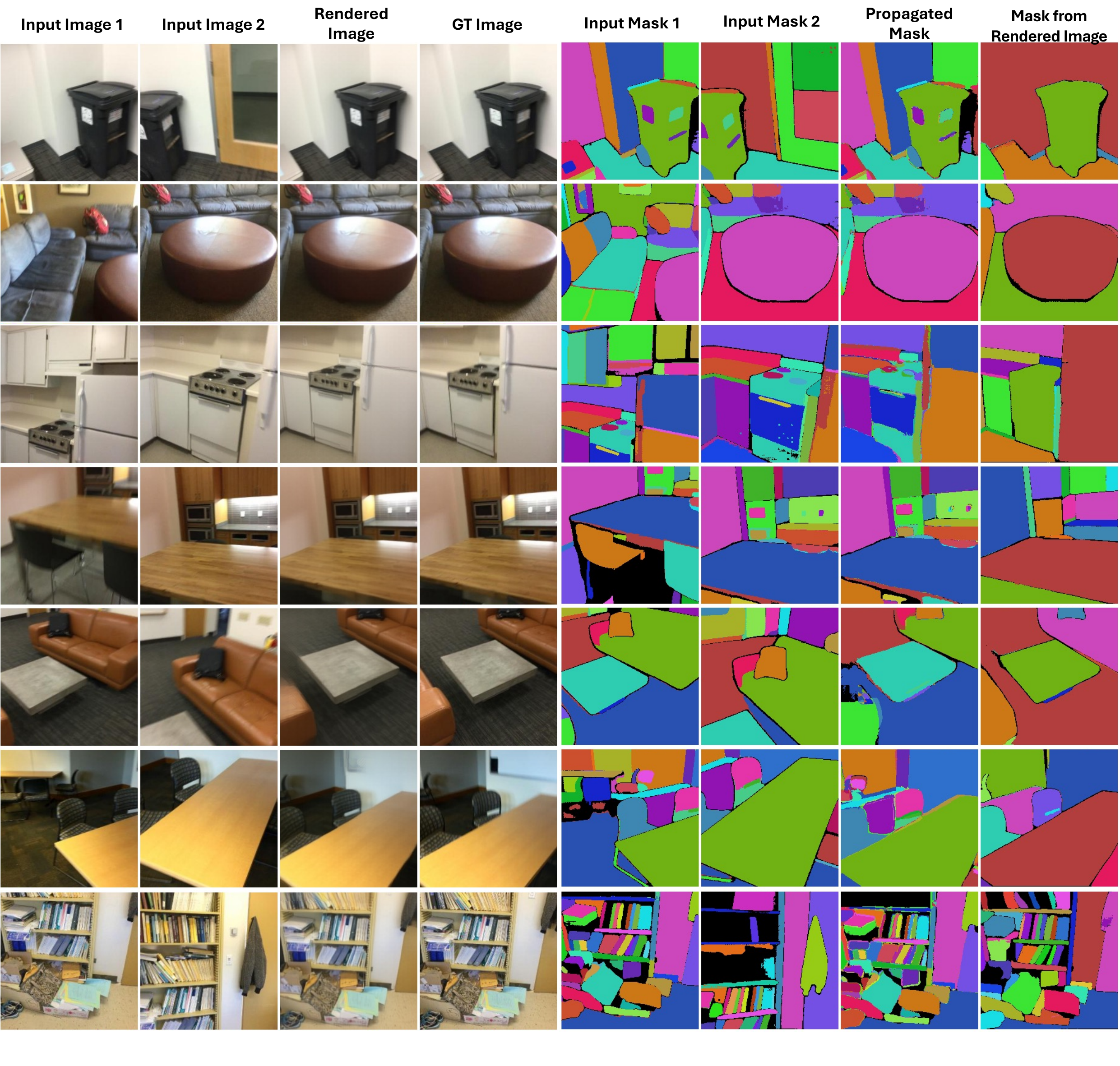}
    \caption{\textbf{Additional qualitative for SAM2 propagation.} Our method propagates SAM2 masks to novel views without additional training.}
    \label{fig:qualitative_supple_sam}
\end{figure}

\subsection{SAM2 Mask Propagation}
Figures~\ref{fig:qualitative_supple_sam} present additional SAM2~\cite{sam2} mask propagation results on ScanNet scenes, showing that SAM2 masks can be used as source-view labels for novel-view propagation. Note that we only run inference on the input views to obtain view-consistent masks.

\subsection{Supplementary Video}
We provide a video file, \texttt{supplement.mp4}, to showcase (1) Panoptic Propagation via Novel View Synthesis model, (2) SAM2~\cite{sam2} Mask propagation on ScanNet dataset~\cite{scannet}. We render video sequences by mapping GT poses and Latent poses into our pipeline. Since Latent Pl\"ucker~\cite{less3depend} encodes camera poses in a learned latent space without ground-truth pose supervision,
the latent pose space does not directly correspond to the metric camera extrinsic space. To obtain smooth and controllable trajectories, we apply a lightweight test-time pose optimization. For each scene, we fit a small residual MLP to map GT poses to the corresponding latent poses using the available key frame views, then interpolate in the ground-truth pose space and convert each intermediate pose to the latent space via the learned mapping. This allows us to render temporally coherent videos that faithfully follow the intended camera path.

\clearpage
\bibliographystyle{splncs04}
\bibliography{main}

\end{document}